\documentclass{article}

\usepackage{microtype}
\usepackage{graphicx}
\usepackage{subfigure}
\usepackage{booktabs} 
\usepackage{tcolorbox}
\tcbuselibrary{breakable}

\usepackage{hyperref}

\usepackage[accepted]{icml2025}

\usepackage{amsmath}
\usepackage{amssymb}
\usepackage{mathtools}
\usepackage{amsthm}

\usepackage{multirow}
\usepackage{xcolor}
\usepackage{colortbl}

\usepackage[capitalize,noabbrev]{cleveref}

\theoremstyle{plain}

\theoremstyle{definition}

\theoremstyle{remark}

\usepackage[textsize=tiny]{todonotes}

\usepackage{enumitem}
\usepackage{color}

\newcommand{\grayline}{\rowcolor[gray]{.90}}

\icmltitlerunning{Reasoning with Reinforced Functional Token Tuning}

\begin{document}

\twocolumn[
\icmltitle{Reasoning with Reinforced Functional Token Tuning}


\icmlsetsymbol{equal}{*}

\begin{icmlauthorlist}
\icmlauthor{Kongcheng Zhang}{zju,comp}
\icmlauthor{Qi Yao}{comp}
\icmlauthor{Baisheng Lai}{comp}
\icmlauthor{Jiaxing Huang}{ntu}
\icmlauthor{Wenkai Fang}{zju}\\
\icmlauthor{Dacheng Tao}{ntu}
\icmlauthor{Mingli Song}{zju}
\icmlauthor{Shunyu Liu}{ntu}
\end{icmlauthorlist}

\icmlaffiliation{zju}{Zhejiang University}
\icmlaffiliation{comp}{Alibaba Cloud Computing}
\icmlaffiliation{ntu}{Nanyang Technological University}

\icmlcorrespondingauthor{Shunyu Liu}{shunyu.liu@ntu.edu.sg}

\icmlkeywords{LLMs}

\vskip 0.3in
]



\printAffiliationsAndNotice{} 

\begin{abstract}
In this work, we propose \textit{\textbf{R}einforced \textbf{F}unctional \textbf{T}oken \textbf{T}uning}~(RFTT), a novel reinforced fine-tuning framework that empowers Large Language Models~(LLMs) with self-play learn-to-reason capabilities. 
Unlike prior prompt-driven reasoning efforts, RFTT embeds a rich set of learnable functional tokens (\textit{e.g.}, \texttt{<analyze>}, \texttt{<verify>}, \texttt{<refine>}) directly into the model vocabulary, enabling chain-of-thought construction with diverse human-like reasoning behaviors.
Specifically, RFTT comprises two phases: (1) supervised fine-tuning performs prompt-driven tree search to obtain \textit{self-generated} training data annotated with functional tokens, which warms up the model to learn these tokens for reasoning; and (2) online reinforcement learning further allows the model to explore different reasoning pathways through functional token sampling without relying on prompts, thereby facilitating effective \textit{self-improvement} for functional reasoning.
Extensive experiments demonstrate the superiority of the proposed RFTT on mathematical benchmarks, significantly boosting Qwen-2.5-7B-Instruct (70.6\% $\to$ 79.8\%) and
LLaMA-3.1-8B-Instruct (32.2\% $\to$ 60.2\%) on the MATH dataset. Moreover, the performance of RFTT consistently improves with more search rollouts at inference time. 
Our code is available at \url{https://github.com/sastpg/RFTT}.

\end{abstract}

\section{Introduction}
Recent advances in Large Language Models~(LLMs), particularly exemplified by OpenAI-o1~\citep{jaech2024openai} and DeepSeek-R1~\citep{guo2025deepseek}, have demonstrated sophisticated reasoning capabilities with remarkable success across various professional domains, such as mathematical analysis~\citep{hendrycks2021measuring,cobbe2021training,patel2021nlp,he2024olympiadbench} and algorithmic programming~\citep{chen2021evaluating,austin2021program,zhuo2024bigcodebench}. Despite the encouraging results achieved, learning to reason remains a crucial yet challenging task for LLMs as it necessitates high-quality reasoning data for training~\citep{xu2025towards}, especially for smaller LLMs with 7B or 8B parameters~\citep{guan2025rstar}. Thus, previous attempts often rely on stronger models or human annotators to generate high-quality Chain-of-Thought~(CoT) data~\citep{min2024imitateexploreselfimprovereproduction,lightman2023let,huang2024o1,wang2023mathcoder}; however, such approaches inevitably incur substantial costs~\citep{wang2024math,guan2025rstar} and are prone to limited scalability~\citep{li2023coannotating,ahn2024large}.

To address these limitations, existing studies explore self-play mechanisms where LLMs can iteratively refine their reasoning capabilities through self-generated rationales without human-curated training data~\citep{hao2023reasoninglanguagemodelplanning,zhang2024rest,wu2024self,zhang2024accessing}.
For instance, techniques such as CoT prompting enable models to generate multi-step reasoning paths for unlabeled questions, followed by rejection sampling to select high-confidence solutions as training data~\citep{yuan2023scaling,rso2024liu,brown2024large}.
Drawing inspiration from AlphaGo~\citep{silver2016mastering}, advanced self-play methods further integrate Monte Carlo Tree Search~(MCTS) to find high-quality reasoning paths by constructing reasoning trees~\citep{hao2023reasoninglanguagemodelplanning,zhang2024rest,zhang2024accessing,guan2025rstar}.
However, these direct tree-search approaches face two fundamental challenges: (1) LLMs tend to generate homogeneous reasoning paths due to their inherent preference for syntactic patterns established during training~\citep{wei2022chain,wang2022self,patil2025advancingreasoninglargelanguage}; (2) the high-dimensional search space further limits the exploration capabilities of LLMs, as the combinatorial vocabulary space leads to an exponential growth in candidate reasoning paths~\citep{zhang2024rest}.

Therefore, a recent work, rStar~\citep{qi2024mutualreasoningmakessmaller}, introduces functional prompts~(\textit{e.g.}, ``propose a one-step thought'', ``rephrase the question'') that steer tree search by simulating human-like reasoning behaviors, thereby diversifying node exploration while constraining the search space. However, rStar operates purely as an inference-time augmentation without internalizing the reasoning capabilities through model training. The functional prompts act as external constraints rather than learned patterns. As a result, each inference of rStar must rely on iterative tree-search traversal across different functional prompts to seek correct solutions. Compared to direct reasoning methods~(which directly generate CoT responses without relying on search), this exhaustive searching approach results in a substantial number of redundant LLM inference calls.

In this work, we propose \textit{\textbf{R}einforced \textbf{F}unctional \textbf{T}oken \textbf{T}uning}~(RFTT), a novel reinforced fine-tuning framework that introduces learnable functional tokens (\textit{e.g.}, \texttt{<analyze>}, \texttt{<verify>}, \texttt{<refine>}) into the model vocabulary to facilitate self-play learn-to-reason. 
Technically, RFTT operates in two phases: 
(1) during the Supervised Fine-Tuning~(SFT) warmup phase, RFTT employs functional prompt-guided MCTS to construct reasoning trees, where both correct and incorrect solution paths are merged to synthesize human-like reasoning paths. These self-generated paths explicitly connect reasoning nodes through functional tokens, thereby forming structured training data for learning to reason with these tokens. 
(2) In the online Reinforcement Learning~(RL) phase, the model transitions from prompt-guided to token-guided reasoning by sampling functional tokens to autonomously expand reasoning trees. Through self-play exploration, the model reinforces high-value reasoning pathways, ultimately achieving autonomous self-improvement of functional reasoning capabilities.
Our core contributions are summarized as follows:
\begin{itemize}[leftmargin=*, itemsep=2pt,topsep=0pt,parsep=0pt]
\item We propose a new learn-to-reason paradigm that pioneers the integration of learnable functional tokens into the model vocabulary, enabling the model to establish internalized token-guided reasoning patterns rather than relying on external prompt-guided constraints.
\item We devise RFTT, a novel reinforced fine-tuning framework for self-play learning to reason. The SFT phase bootstraps reasoning through \textit{self-generated} training data annotated with functional tokens, while the RL phase enables autonomous exploration of reasoning pathways through token-guided tree traversal, achieving \textit{self-improvement} for functional reasoning.
\item Extensive experiments conducted on various mathematical benchmarks demonstrate that RFTT yields significantly superior results to state-of-the-art counterparts. Notably, the performance of RFTT continues to improve as the number of search rollouts increases during inference.
\end{itemize}

\begin{figure*}[t]
    \centering
    \includegraphics[width=\linewidth]{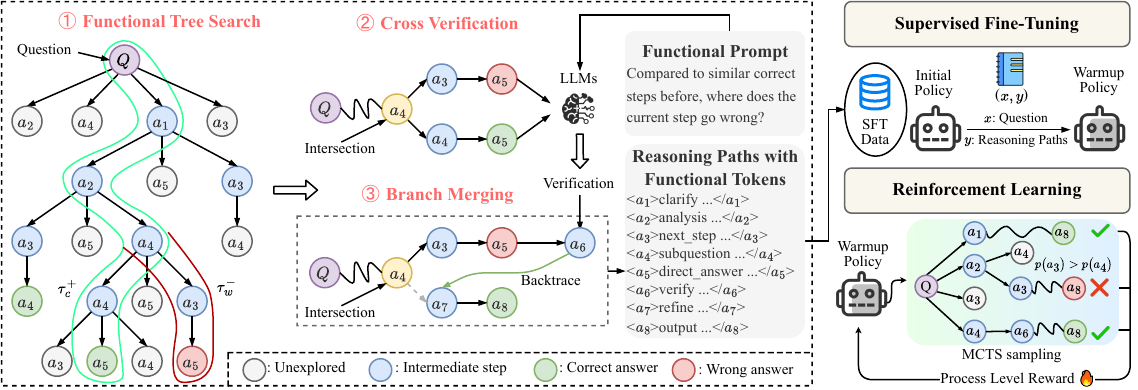}
    \vspace{-18pt}
    \caption{(Left) A conceptual illustration of reasoning path generation based on functional tree search.
    (Right) An overview of the RFTT framework that comprises two phases: supervised fine-tuning warmups the model with self-generated functional token-annotated data, while online reinforcement learning allows the model to perform autonomous reasoning path exploration for self-improvement.
    }
    \label{fig:framework}
    \vspace{-8pt}
\end{figure*}

\section{Related Works}
\textbf{Mathematical Reasoning.}
Recent advancements in LLMs for mathematical reasoning have primarily focused on two broad categories of approaches: prompt engineering and learning to reason. Prompt engineering techniques design specific prompting strategies, such as CoT~\citep{wei2022chain,kojima2022large,wang2022self,zhang2022automatic}, multi-path reasoning~\citep{yao2023tot,yao2022react,besta2024graph} or divide-and-conquer~\citep{zhou2022least,sel2023algorithm,wang2023plan} to guide the model to generate step-by-step reasoning processes during the inference phase. In contrast, learn-to-reason approaches train the model explicitly to improve its intrinsic reasoning ability. Extensive math and code corpora are first utilized to build the fundamental reasoning capability in the pre-training phase~\citep{lewkowycz2022solving,shao2024deepseekmath,yang2024qwen2,abdin2024phi,dubey2024llama} and various techniques, such as supervised fine-tuning~\citep{yuan2023scaling,wang2023mathcoder,min2024imitateexploreselfimprovereproduction}, direct preference optimization~\citep{rafailov2024direct,lai2024step} and reinforcement learning~\citep{reft2024,shao2024deepseekmath,guo2025deepseek} are used to further enhance the mathematical reasoning ability during post-training.

\textbf{Tree-Search Reasoning.}
While most CoT methods have typically adopted an auto-regressive reasoning manner, there has been a trend to engage in more complicated reasoning architectures like trees~\citep{koh2024tree,zhang2024llamaberrypairwiseoptimizationo1like}. Recently, various methods for exploring tree structures have been devised to identify optimal reasoning trajectories, \textit{e.g.},  tree of thought~\citep{yao2023tot}, graph of thoughts~\citep{besta2024graph} and Monte Carlo tree search~\citep{hao2023reasoninglanguagemodelplanning,chen2024alphamath,zhang2024rest}. These direct tree-search methods struggle with limited exploration due to the homogeneous reasoning paths and the large search space. Although rStar~\citep{qi2024mutualreasoningmakessmaller} utilizes functional prompts in tree search to diversify and constrain the search process, they only act as external constraints in the inference phase while we introduce learnable functional tokens during model training to facilitate effective and efficient exploration, thus improving the model's intrinsic reasoning ability.

\textbf{LLM Self-Improvement.}
Self-improvement with self-generated reasoning paths has gained increasing interest in mathematical reasoning due to the lack of golden rationales. Early methods~\citep{yuan2023scaling,zelikman2022star} utilize reject sampling to select reasoning paths with correct answers for iterative self-training, which often leads to local optimal due to the homogeneity of self-generated reasoning paths. Recent approaches explore MCTS~\citep{zhang2024rest,tian2024toward,hao2023reasoninglanguagemodelplanning,zhang2024accessing} to generate more diverse reasoning paths offline or adopt random sampling to produce abundant reasoning paths during the online reinforcement learning phase~\citep{reft2024,wang2024math,lambert2024t}. However, these methods struggle to fully explore the reasoning space, especially for smaller language models. In RFTT, we employ function tokens for automatic exploration of the reasoning space through token-guided tree traversal, achieving self-improvement for functional reasoning.

\begin{figure*}[t]
    \centering
    \includegraphics[width=\linewidth]{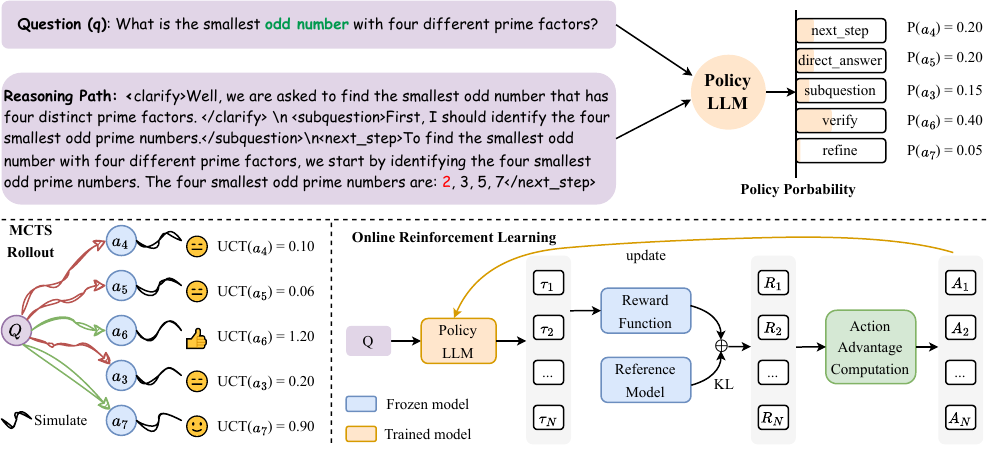}
    \vspace{-18pt}
    \caption{An illustrative diagram of the online reinforcement learning phase in RFTT. The LLM policy directly samples functional tokens from its vocabulary to autonomously expand reasoning trees to search for the final solutions. Then we use online reinforcement learning with process rewards to optimize the functional reasoning capabilities of the LLM policy.
    }
    \label{fig:rftt}
    \vspace{-8pt}
\end{figure*}

\section{Methodology}
In this section, we first provide the problem formulation. Then we further detail the proposed learn-to-reason framework of RFTT. The whole procedure is outlined in Fig.~\ref{fig:framework}.

\subsection{Problem Formulation}
We consider a complex reasoning task as a multi-step reasoning generation problem, which decomposes the problem into a sequence of simpler intermediate steps through structured reasoning operations. Assuming a problem \(x\) requires \(T\) intermediate reasoning steps, the reasoning path can be denoted as $\tau=\{s_{0}, s_{1}, s_{2}, ..., s_{T}\}$, where $s_0 =x$ and $s_t$ is the $t$-th intermediate reasoning steps. Furthermore, we model this process as a tree search, where the root node represents the problem \(x\), the edges denote the action space $\mathcal{A}$ (which can be special tokens or prompts), and children nodes are next steps generated by a policy model \(\pi_\theta\) with parameter $\theta$ under corresponding actions. Each distinct path from the root to the terminal node denotes a sequence of decision-making steps that constitutes a potential solution trajectory $\tau$. For any given problem \(x\), the subset of its solution space $\mathcal{T} = \{\tau_{1}, \tau_{2}, ..., \tau_{N}\}$ can be extracted from the tree structure. Our aim is to sample diverse reasoning paths and ultimately derive optimal solutions.

Considering the problem setup above, we can transfer the task of better reasoning to better selecting actions based on previous steps. In this work, we focus on how to inspire intrinsic reasoning abilities with functional tree search. The conventional natural language generation task often regards vocabulary-level token sampling as actions. If we adopt vocabulary-level actions to conduct tree search on LLMs, the action space can be extremely large, leading to a sharp increase in computational complexity and reduced search efficiency. Thus we design a series of human-like reasoning behaviors as action space. During the preliminary warmup phase for SFT, we employ functional prompts as the action space for exploratory tree search, while transitioning to functional tokens during RL optimization.

\subsection{Functional Tokens with Prompts}
We propose several functional tokens with prompts as actions to emulate how humans engage in complex reasoning:
\begin{itemize}[leftmargin=*, itemsep=1pt,topsep=0pt,parsep=0pt]
    \item \texttt{\textbf{<clarify>}} (\(a_1\)): Organize the conditions of the given complex question and express them in a clearer form to avoid misunderstanding.
    \item \texttt{\textbf{<analysis>}} (\(a_2\)): Analyze the general problem-solving idea and the knowledge that might be involved.
    \item \texttt{\textbf{<subquestion>}} (\(a_3\)): Break down complex reasoning problems into sub-problems that are easier to tackle.
    \item \texttt{\textbf{<next\_step>}} (\(a_4\)): Propose the next intermediate step based on the existing reasoning steps.
    \item \texttt{\textbf{<direct\_answer>}} (\(a_5\)): Complete the remaining steps until achieving the final answer step-by-step.
    \item \texttt{\textbf{<verify>}} (\(a_6\)): Reflect if there are any mistakes in the preceding reasoning steps considering LLMs might make computational or logical errors in a reasoning step.
    \item \texttt{\textbf{<refine>}} (\(a_7\)): Correct certain previous steps if the errors are identified, ensuring that each step is accurate to achieve a final outcome.
    \item \texttt{\textbf{<output>}} (\(a_8\)): Output the final answer that is in the corresponding expected format.
\end{itemize}
All functional prompts are detailed in Appendix~\ref{app:prompt}.

\subsection{Functional Monte Carlo Tree Search}
\label{sec:3.2}
The trajectories generated by direct tree search methods are often homogeneous due to the inherent preference for syntactic patterns in LLMs. To address this, we introduce MCTS guided by functional prompts in the SFT phase (or functional tokens in the RL phase) as actions to obtain diverse reasoning nodes. Concretely, we use functional prompts to search and adopt branch merging to connect intermediate nodes using functional tokens, forming structured reasoning paths for SFT. During RL, we employ functional tokens learned in SFT to explore different reasoning paths without relying on prompts.

\textbf{Step 1: Functional Tree Search}

We now proceed to details on how to employ functional prompts in SFT (or functional tokens in RL) to search candidate reasoning paths, which consists of four basic operations: \textit{selection}, \textit{expansion}, \textit{simulation} and \textit{backpropogation}.

Starting from root node (question) \(s_0=x\), a series of internal nodes $\{s_1, s_2, ..., s_l\}$ are chosen until reaching a leaf node $s_l$. In the selection phase, we use the Upper Confidence Bound applied to trees~(UCT) to select each node for balancing exploration and exploitation:
\begin{equation}
    \texttt{UCT}(s) = \frac{Q(s)}{N(s)} + c\cdot\sqrt{\frac{\ln \tilde{N}(s) }{N(s)} }, \label{eq2}
\end{equation}
where $Q(s)$ is the value of node $s$, $N(s)$ is the number of times visited $s$, $\tilde{N}(s)$ is the number of times visited the parent node of $s$, and $c$ is the exploration weight.

In the expansion phase, the selected leaf node $s_l$ is expanded by action space $\mathcal{A} = \{a_1, a_2, a_3, a_4, a_5, a_8\}$. We use action $a_i \in \mathcal{A}$ (functional prompt in SFT or functional token in RL) to guide the LLM to generate the next reasoning step. The simulation phase iteratively selects and expands from node $s_l$ until reaching a terminal node, which we define as achieving either the final answer or the maximum permitted tree depth. Once the MCTS reaches a terminal node $s_T$, a backpropagation is carried out. During this phase, the outcome of a trajectory is propagated back up the tree to update the node information such as visited time and node value $Q$. More details of MCTS are presented in Appendix~\ref{app:mcts}.

We call the above process a rollout, where we can extract a trajectory $\tau = \{s_0, s_1, ..., s_T\}$ from the root node \(s_0=x\) to terminal node \(s_T\). After MCTS rollout iterations, we can extract multiple reasoning trajectories $\mathcal{T} = \{\tau_1, \tau_2, ..., \tau_N\}$ for each question from a tree structure. The following steps describe how we utilize them to construct structured human-like reasoning paths for SFT.

\textbf{Step 2: Cross Verification}

To construct a complex reasoning structure that incorporates self-verification for SFT, we generate an extra node via cross verification between two trajectories and thereby seamlessly concatenate them. Given a question \(x\) and its ground truth \(y\), we can get candidate reasoning trajectories $\mathcal{T}$ as stated in step 1. We first choose a correct trajectory with the highest average process rewards given by a process reward model:
\begin{equation}
\tau_c = \mathop{\arg\max}_{\tau_i \in \mathcal{T}_c} \bar{R}(\tau_i),
\end{equation}
where $\mathcal{T}_c=\{\tau_i \in  \mathcal{T} \mid \texttt{ANS}(\tau_i) =  y\}$, \texttt{ANS$(\tau_i)$} is the final answer of trajectory $\tau_i$, and $\bar{R}(\tau_i)$ represents the average rewards of all intermediate steps in trajectory $\tau_i$. We can further separate a subset $\mathcal{T}_w$ from $\mathcal{T}$, which contains the wrong trajectories that intersects with $\tau_c$:
\begin{equation}
\mathcal{T}_w  = \{\tau_i \in  \mathcal{T} \mid  d(\tau_i, \tau_c) \textless \alpha ~\text{and}~ \texttt{ANS}(\tau_i) \neq  y\},
\end{equation}
where $d(\tau_i, \tau_c)$ denotes the number of distinct steps between $\tau_i$ and $\tau_c$, and $\alpha$ is a predetermined hyperparameter.
Here we can derive a reasoning path with a wrong answer:
\begin{equation}
\tau_w = \mathop{\arg\min}_{\tau_i \in \mathcal{T}_w} \bar{R}(\tau_i).
\end{equation}
If $\tau_w \neq \emptyset$, we can represent the set of intersecting nodes of $\tau_c$ and $\tau_w$ as $\tau^+ = \tau_c \cap \tau_w$, and the set of remaining nodes of them as ${\tau_c}^- = \tau_c - \tau^+$ and $\tau^-_w = \tau_w - \tau^+$. Then, we can verify why an error occurs in $\tau^-_w$ using a prompt $p$:
\begin{equation}
s_v \sim \pi_\theta\left( \cdot \mid \tau_c \cup \tau^-_w \cup \{p\}\right),
\end{equation}
where $p$ = ``Compared to similar correct steps before, where does the current step go wrong?''.
Through this approach, we can yield a node $s_v$ with self-verification.

\textbf{Step 3: Branch Merging}

We adopt branch merging to form structured reasoning paths that feature self-verification and self-correction:
\begin{equation}
    \tau_f = \tau^+ \cup \tau^-_w \cup \{s_v\} \cup {\tau_c}^-.
\end{equation}
Finally, we concatenate intermediate reasoning steps in trajectories by functional tokens, generating a trainable reasoning path for warmup in supervised fine-tuning:
\begin{equation}
\mathcal{D}_{\text{SFT}} = \left\{ \otimes_{s_i \in \tau_f} \left( \texttt{<}a_i\texttt{>}s_i\texttt{<}/a_i\texttt{>} \right) \right\},
\end{equation}
where $\otimes$ denotes the string concatenation operation, $s_i$ is the intermediate reasoning step, and $a_i$ is the corresponding functional token.

\subsection{Reinforced Functional Token Tuning}
Our primary goal is to empower LLMs with self-play learn-to-reason capabilities. To achieve this, we introduce a two-phase training strategy: the SFT phase learns functional tokens for reasoning by self-generated training data annotated with these tokens, while the RL phase further enables autonomous exploration of reasoning paths through functional token-guided tree search, thereby reinforcing high-value reasoning paths.

\textbf{Phase 1: Supervised Fine-Tuning}

In this stage, an initial base model $\pi_\theta$ is warmuped on the dataset $\mathcal{D}_{\text{SFT}}$ that consists of the tuples of ``(question, reasoning path)'' to obtain SFT model $\pi_{\text{SFT}}$. Specifically, we treat functional tokens as new special tokens that can be embedded directly into the model vocabulary. 
During the SFT phase, the model learns to reason with these functional tokens without relying on prompts.

\textbf{Phase 2: Online Reinforcement Learning}

After being trained with functional tokens, the policy model \(\pi_\theta\) can explicitly and dynamically select functional tokens and further generate the specific reasoning steps associated with these tokens. Through this design, the policy model can reinforce the use of key functional token combinations via a form of online self-play, significantly improving its learning efficiency. Specifically, the policy model learns by iteratively searching, evaluating the process in each rollout, and updating its parameters, as illustrated in Fig.~\ref{fig:rftt}. Note that the functional token can significantly reduce the search space of tree search.

Like AlphaZero~\citep{silver2018general}, the functional token is sampled based on the log-likelihood scores if there exist unexplored actions, otherwise select them according to the UCT score:
\begin{equation}
a_t = 
    \begin{cases}
        \text{max}_{a \in \mathcal{A}} ~\pi_\theta(a|s_{0:t}), & \exists ~a \in U(s_{t}), \\
        \text{max}_{a \in \mathcal{A}} ~\texttt{UCT}(s_{t}),     & \text{otherwise}, \\
    \end{cases}
\end{equation}
where $U(s_{t})$ denotes unexplored functional tokens of intermediate reasoning step $s_{t}$ and $s_{0:t}$ represents all intermediate steps from timestep $0$ to $t$.

Furthermore, the reward function is defined as:
\begin{equation}
    R_t(s_{0:t}, a_t, s_{t+1}) = \texttt{RM}(s_{0:t}, a_t, s_{t+1}) - \beta \cdot \texttt{KL}(t),
\end{equation}
\begin{equation}
    \texttt{RM}(s_{0:t}, a_t, s_{t+1})\! =\! 
        \begin{cases}
            1, &  \!\!\!\!\! \texttt{ANS}(s_{t+1}) = y, \\
            0.1, &\!\!\!\!\! \texttt{ANS}(s_{t+1}) \neq \texttt{null}, \neq y, \\
            \sigma, & \!\!\!\!\! \texttt{ANS}(s_{t+1}) = \texttt{null},
    \end{cases}
\end{equation}
\begin{equation}
    \texttt{KL}(t) = \log\left(\frac{\pi_{\theta}(a_t|s_{0:t})}{\pi_{\text{SFT}}(a_t|s_{0:t})}\right),
\end{equation}
where the SFT model $\pi_{\text{SFT}}$ serves as the reference model, $\beta$ is KL coefficient , and $\sigma$ is the process reward. Note the $\sigma$ can be assigned to $0$ if using outcome rewards. Our approach can be flexibly integrated with a Process Reward Model~(PRM) or an Outcome Reward Model~(ORM).

During online RL, the model parameters are optimized using the Reinforce++ algorithm~\cite{hu2025reinforce++}. The core policy objective combines clipped updates with normalized advantage values through the following loss function:
\begin{multline}
\mathcal{L}_{RL}(\theta)=\mathbb{E}_{t}\left[\min \left(\!\frac{\pi_\theta(a_t | s_{0:t})}{ \pi_{\theta_{\text{old}}}(a_t | s_{0:t})} \hat{A}_{t},\right.\right.\\\left.\left.\text{clip}\left(\frac{\pi_\theta(a_t | s_{0:t})}{ \pi_{\theta_{\text{old}}}(a_t | s_{0:t})}, 1 - \epsilon, 1  + \epsilon\right)\hat{A}_{t}\!\right)  \right],
\end{multline}
where $\pi_{\theta_{\text{old}}}$ is the previous policy model, $\epsilon$ is the clipping coefficient, and the normalized advantage value $\hat{A}_{t}$ is calculated based on the reward $R_t$.

Overall, the transition from prompt-guided to token-guided reasoning enables the effective and efficient exploration of the reasoning paths during the RL phase. The policy model learns to reinforce high-value reasoning paths through functional token-guided tree search and achieves self-improvement for functional reasoning.

\begin{table*}[!t]
\caption{Accuracy of our proposed RFTT and baselines across different mathematical reasoning benchmarks. The best results in each box are highlighted in \textbf{bold}. The proposed RFTT significantly boosts the performance of smaller LLMs across all datasets.}
\label{tab:main_results}
\vspace{-3pt}
\begin{center}
\begin{small}
\begin{tabular*}{\textwidth}{@{\extracolsep{\fill}}llcccccc}
\toprule
\multirow{2}{*}{\textbf{Model}}  & \multirow{2}{*}{\textbf{Setting}} & \textbf{In-Domain} & \multicolumn{4}{c}{\textbf{Out-of-Domain}} & \multirow{2}{*}{\textbf{AVERAGE}} \\
\cmidrule(lr){3-3} \cmidrule(lr){4-7}
 &  & MATH & GSM8K & SVAMP & Olympiad Bench & AMC & \\
\midrule
\grayline \multicolumn{8}{c}{\textit{Superior LLMs}}\\
\midrule
GPT-4                                 & Closed-source  & 64.5 & 94.2 & 92.6   & 32.0   & 42.5 & 65.16\\
GPT-4o                                & Closed-source  & 76.6 & 96.1 & \textbf{93.8}  & \textbf{43.3}   & 47.5 & 71.46\\
GPT-4o mini                           & Closed-source  & 70.2 & 93.2 &  89.2 & 35.8   & 55.0 & 68.68\\
o1-preview                            & Closed-source& \textbf{85.5} & 93.0 & -  & -   & \textbf{90.0} & -\\
Claude-3.5-Sonnet                     & Closed-source  & 71.1 & \textbf{96.4} & -  & -   & - & -\\
LLaMA-3.1-70B-Instruct                & Open-source 70B& 68.0 & 95.1 & -  & 27.7   & 30.0 & -\\
Qwen-2.5-14B-Instruct                  & Open-source 14B& 80.0 & 93.2 & 91.6  & 39.4  & 52.5  & 71.34\\
\midrule
\grayline \multicolumn{8}{c}{\textit{Small LLMs}}\\
\midrule
\multirow{7}{*}{Qwen-2.5-7B-Instruct}  & Zero-shot      & 70.6 & 90.0 & 84.9  & 33.8 & 40.0 & 63.86\\
                                      & Zero-shot CoT  & 72.0 & 91.1  & 88.5 & 35.1   & 45.0 & 66.34 \\
                                      & Few-shot CoT   & 75.5 & 91.6 & 90.2  & 36.5   & 45.0 & 67.76 \\
                                      & CoT+SC@4       & 76.4 & 92.1 & 90.6  & 38.2   & 50.0 & 69.46\\
                                      & ReFT           & 75.8 & 94.1 & 89.3  & 38.1   & 60.0 & 71.46\\
                                      & SFT Warmup (Ours)    & 73.2 & 91.9 & 87.4 & 36.8   & 60.0 & 69.86\\
                                      & RFTT (Ours)    & \textbf{79.8} & \textbf{95.2} & \textbf{93.0}  & \textbf{40.3}   & \textbf{70.0} & \textbf{75.66}\\
\midrule
\multirow{7}{*}{LLaMA-3.1-8B-Instruct}& Zero-shot      & 32.2 & 76.6 & 68.3  & 15.4   & 15.0 & 41.50\\
                                      & Zero-shot CoT  & 50.6 & 84.5 & 78.2  & 18.6   & 17.5 & 49.88\\
                                      & Few-shot CoT   & 53.0 & 85.1 & 82.0  & 20.0   & 20.0 & 52.20\\
                                      & CoT+SC@4       & 52.2 & 86.0 & 85.6  & 19.7   & 25.0 & 53.70\\
                                      & ReFT    & 55.0 & 90.8 & 84.3  & 24.9 & 42.5 & 59.50 \\
                                      & SFT Warmup (Ours)    & 53.2 & 86.4 & 82.9  & 24.6  & 40.0 & 57.40\\
                                      & RFTT (Ours)    & \textbf{60.2} & \textbf{91.9} & \textbf{87.5}  &\textbf{29.8}   & \textbf{55.0} & \textbf{64.88}\\
\midrule
\multirow{7}{*}{LLaMA-3.2-3B-Instruct}& Zero-shot      & 32.4 & 69.6 & 68.2  & 14.6   & 12.5 & 39.46\\
                                      & Zero-shot CoT  & 46.8 & 75.8 & 78.1  & 15.7   & 17.5 & 46.78\\
                                      & Few-shot CoT   & 47.2 & 77.7 & 77.3  & 17.4   & 17.5 & 47.42\\
                                      & CoT+SC@4       & 49.8 & 79.5 & 80.5  & 19.6   & 22.5 & 50.38\\
                                      & ReFT           & 52.4 & 78.5 & 81.6  & 22.8 & 35.0 & 54.06 \\
                                      & SFT Warmup (Ours)    & 50.6 & 77.4 & 79.1  & 20.6  & 35.0 & 52.54\\
                                      & RFTT (Ours)    & \textbf{56.2} & \textbf{82.9} & \textbf{84.2}  &\textbf{24.5}   & \textbf{47.5} & \textbf{59.06}\\
\bottomrule
\end{tabular*}
\end{small}
\end{center}
\vspace{-20pt}
\end{table*}

\section{Experiments}
\subsection{Experimental Setups}
\label{sec:4.1}
\textbf{Datasets.} 
For training, we only adopt the training set of the MATH dataset as our training data.
For evaluation, we employ five established mathematical reasoning benchmarks: MATH~\citep{hendrycks2021measuring}, GSM8K~\citep{cobbe2021training}, SVAMP~\citep{patel2021nlp}, Olympiad Bench~\citep{he2024olympiadbench},  and AMC\footnote{https://huggingface.co/datasets/AI-MO/aimo-validation-amc}. To ensure evaluation efficiency and prevent data contamination, we use the MATH-500~\citep{lightman2023let} for MATH evaluations while retaining full test sets for other datasets. More information about our training and evaluation dataset are listed in Appendix~\ref{app:exp}.

\textbf{Base Models.} RFTT is generally applicable to a wide range of LLMs. In our experiments, we select three popular open-source LLMs: LLaMA-3.2-3B-Instruct, LLaMA-3.1-8B-Instruct~\citep{dubey2024llama} and Qwen-2.5-7B-Instruct~\citep{yang2024qwen2}. By focusing on LLMs with relatively small parameters under 10B, we expect that RFTT enables the smaller model to learn to reason via self-play, ultimately achieving results comparable to or exceeding larger LLMs in complex reasoning tasks.

\textbf{Baselines.} We compared RFTT against four categories of baselines in our experiments: (1) \textit{Superior LLMs}, including leading closed-source models and open-source models;  (2) \textit{Model fine-tuning methods}, including SFT and ReFT~\citep{reft2024}; (3) \textit{In-context learning methods}, including Zero-shot, Zero-shot CoT, Few-shot CoT, and CoT+SC@4; (4) \textit{Tree search methods}, including ResT-MCTS*~\citep{zhang2024rest}, rStar~\citep{qi2024mutualreasoningmakessmaller} and LLaMA-Berry~\citep{zhang2024llamaberrypairwiseoptimizationo1like}. 

\textbf{Evaluation Metric.} We employ Pass@1 accuracy across all benchmark datasets as our evaluation metric, with correctness determined through comparison between generated outputs and ground truth. To extract answers reliably, we require the model to wrap its final answer in ~\texttt{boxed\{\}}. 

\subsection{Implementation Details}
All experiments run on 8$\times$A800-80GB GPUs.
While performing MCTS to collect diverse reasoning paths via various functional prompts, we use Qwen-2.5-7B-Instruct as the generator and math-shepherd-mistral-7b-prm~\citep{wang2024math} as the process reward model.
Note that for the Qwen experiment, the model was trained on its own generated data. For convenience, we used this data directly in other experiments without regenerating it. We present the performance results after both SFT and RL, and the findings are consistent.
We set the maximum search depth to 15 and perform 16 rollouts per question. For sampling each step, we employ the vLLM engine~\citep{kwon2023efficient} to accelerate with the temperature set to 0.9, top p set to 0.8, and max tokens set to 1024. We spend approximately one day searching through 1.2K questions using 64 concurrent processes, and then gather 1K SFT data as described in Sec.~\ref{sec:3.2}.

We use LLaMA-Factory~\cite{zheng2024llamafactory} and OpenRLHF~\cite{hu2024openrlhf} for SFT and RL respectively. During SFT, the training batch size is 128, the learning rate is 7e-6, and the cutoff length is 8192. We train the initial model for 10 epochs on about 1k CoT data with functional tokens and select the checkpoint with the best performance as the initial policy model for RL.
During RL, we implement functional token-guided MCTS to sample different reasoning paths for a question. After that, we can choose to score the intermediate reasoning nodes in the tree. At each training step, the model searches for 16 reasoning paths per question with a batch of 16 distinct questions. The learning rate for the policy model is set to 5e-7, the temperature is set to 0.95, and the KL coefficient is set to 0.01.

\subsection{Main Results}
\textbf{Results on Different Benchmarks.} 
We conduct a comprehensive evaluation of the effectiveness of RFTT across different mathematical benchmarks. Table~\ref{tab:main_results} presents a comparative analysis between our framework and state-of-the-art baselines. We highlight three key findings: (1) RFTT can significantly improve the complex reasoning proficiency of small-parameter LLMs. For example, LLaMA-3.1-8B-Instruct initially achieves 50.6\% accuracy on the MATH benchmark using CoT prompting technique. However, after training by RFTT, its Pass@1 accuracy improves to 60.2\%, surpassing self-consistency sampling. Similarly, Qwen-2.5-7B-Instruct with RFTT achieves performance on par with Qwen-2.5-14B-Instruct (79.6\% vs. 80\% on MATH), indicating that smaller models have developed robust reasoning capabilities without any reasoning prompts.
(2) RFTT outperforms ReFT on various mathematical reasoning tasks. We observe that RFTT achieves consistent performance improvements over ReFT, notably surpassing it by an average margin of 5\% across different benchmark evaluations. This result demonstrates the robustness of RFTT and the enormous potential for further self-play training. (3) RFTT has also shown strong generalization ability on other challenging mathematical problems, including AMC and Olympiad Bench. As mentioned in Sec~\ref{sec:4.1}, our training dataset only consists of the training data from MATH, and there may be a risk of over-optimization on well-known testing benchmarks such as MATH and GSM8K. Our results reveal that RFTT remains effective even when applied to unseen problem sets.

\begin{table}[!t]
\caption{Comparison of different tree search methods.}
\label{tab:search_effect}
\vspace{-5pt}
\begin{center}
\begin{small}
\setlength{\tabcolsep}{6pt}
\renewcommand{\arraystretch}{1.2}
\begin{tabular}{@{\extracolsep{\fill}}clcc}
\toprule
\textbf{Dataset} & \textbf{Method} & \textbf{Accuracy} & \textbf{Time per Rollout (s)}\\
\midrule
\multirow{3}{*}{GSM8K}                & rStar    & 92.1\% & 276 \\                     
                                      & LLaMA-Berry & 94.9\% & 339 \\
                                      & Ours  & 95.2\% & 81 \\
\midrule
\multirow{3}{*}{MATH}                 & rStar    & 61.0\% & 526 \\                     
                                      & LLaMA-Berry & 69.4\% & 674 \\
                                      & Ours  & 72.0\% & 131 \\
\bottomrule
\end{tabular}
\end{small}
\end{center}
\vspace{-25pt}
\end{table}

\begin{figure*}[!t]
    \centering
    \includegraphics[width=\linewidth]{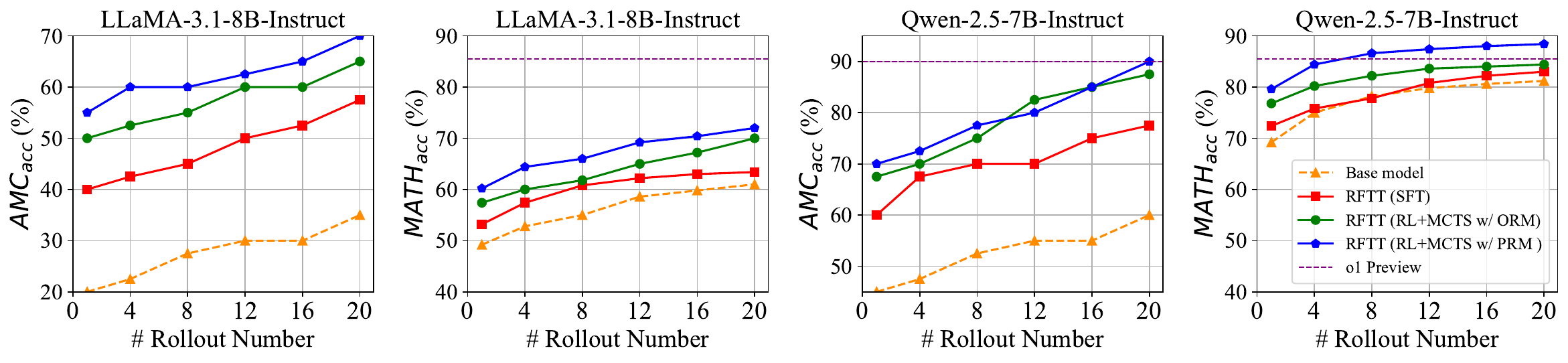}
    \vspace{-25pt}
    \caption{Performance gains under scaling up the inference-time computation on the MATH and AMC benchmarks.}
    \vspace{-5pt}
    \label{fig:scaling_test}
\end{figure*}

\textbf{Scaling Inference-Time Computation.} We integrate learnable functional tokens into the model vocabulary in RFTT, enabling the model to establish internalized token-guided tree search. We use MCTS to augment the policy model trained by RFTT without any guidance of reward models. Fig.~\ref{fig:scaling_test} shows the performance of scaling inference-time computation by comparing the accuracy of policy models with different training strategies across different rollout numbers. The accuracy of one rollout corresponds to the Pass@1 accuracy of the policy model. We highlight two key observations: (1) Scaling inference-time computation enhances mathematical reasoning across all benchmarks, albeit with distinct trends observed. On MATH, the policy models demonstrate a slow improvement after 16 rollouts, whereas on AMC, the accuracies continue to improve steadily. (2) With 8 and 20 rollouts on MATH and AMC benchmarks respectively, the performance of the trained policy model exceeds o1-preview, demonstrating its effectiveness.

\begin{table}[!t]
\caption{Performance gains under self-improvement.}
\label{tab:self_improve}
\vspace{1pt}
\begin{center}
\begin{small}
\setlength{\tabcolsep}{6pt}
\renewcommand{\arraystretch}{1.2}
\begin{tabular}{@{\extracolsep{\fill}}clccc}
\toprule
\textbf{Dataset} & \textbf{Method} & \textbf{1st Iter} & \textbf{2nd Iter} & \textbf{Ratio} $\uparrow$ \\
\midrule
\multirow{2}{*}{GSM8K} & ReST-MCTS*  & 85.0 & 86.2 & 1.4\% \\                     
                      & Ours    & 86.4 &  88.1 & \textbf{2.0\%} \\
                      \midrule
\multirow{2}{*}{MATH} & ReST-MCTS*  & 47.6 & 48.4 & 1.7\% \\
                      & Ours    & 53.2 & 57.4 & \textbf{7.9\%} \\
\bottomrule
\end{tabular}
\end{small}
\end{center}
\vspace{-25pt}
\end{table}

\begin{table*}[!t]
\caption{Ablation study on different components of RFTT.}
\vspace{2pt}
\label{tab:ablation}
\begin{center}
\begin{small}
\begin{tabular*}{\textwidth}{@{\extracolsep{\fill}}llcccccc}
\toprule
\textbf{Model}                              & \textbf{Setting}  & \textbf{MATH}& \textbf{GSM8K} & \textbf{SVAMP}   & \textbf{Olympiad Bench} & \textbf{AMC} & \textbf{AVERAGE}\\
\toprule
\multirow{4}{*}{Qwen-2.5-7B-Instruct}  & RFTT w/o SFT Warmup      & 74.0 & 87.2 & 90.6  & 33.9   &  47.5 & 66.64\\
                                      & RFTT w/o MCTS  & 75.2 & 93.8  & 91.2 & 39.0   & 62.5 & 72.34\\
                                      & RFTT w/o PRM    & 77.2 & 94.1 & 92.4  & 38.4   & 67.5 & 73.92\\
                                      & RFTT (Ours)    & \textbf{79.8} & \textbf{95.2} & \textbf{93.0}  & \textbf{40.3}   & \textbf{70.0} & \textbf{75.66}\\

\midrule
\multirow{4}{*}{LLaMA-3.1-8B-Instruct}  & RFTT w/o SFT Warmup    & 54.8 & 84.9 & 83.7  & 24.0   & 17.5 & 53.00\\
                                      & RFTT w/o MCTS  & 56.2 & 91.5  & 84.0 & 29.1 & 42.5 & 60.66\\
                                      & RFTT w/o PRM    & 57.4 & \textbf{92.2} & 86.3  & 28.6   & 50.0 & 62.90\\
                                      & RFTT (Ours)    & \textbf{60.2} & 91.9 & \textbf{87.5}  &\textbf{29.8}   & \textbf{55.0} & \textbf{64.88}\\
\midrule
\multirow{4}{*}{LLaMA-3.2-3B-Instruct}  & RFTT w/o SFT Warmup    & 50.6 & 77.2 & 78.8  & 17.2   & 17.5 & 48.26\\
                                      & RFTT w/o MCTS  & 51.8 & 79.2  & 81.2 & 23.9 & 40.0 & 55.22\\
                                      & RFTT w/o PRM   & 54.2 & \textbf{84.1} & 83.6 & \textbf{24.8}   & 42.5 & 57.84\\
                                      & RFTT (Ours)    & \textbf{56.2} & 82.9 & \textbf{84.2}  &24.5   & \textbf{47.5} & \textbf{59.06}\\
\bottomrule
\end{tabular*}
\end{small}
\end{center}
\vspace{-15pt}
\end{table*}

\textbf{Search Efficiency.}
Our proposed RFTT can dynamically select the next high-value functional token based on a partially generated reasoning trajectory, thus enabling a more efficient exploration of the reasoning space. Here, we compare the search efficiency and accuracy of RFTT with other tree-based reasoning methods. Specifically, We measure the average time needed for a single rollout and the proportion of questions with correct search answers across the dataset. As listed in Table~\ref{tab:search_effect}, our framework achieves higher accuracy while substantially reducing time complexity compared to other methods. It outperforms rStar, which relies on extensive node expansion to select a potentially better action, and LLaMA-Berry, which generates a complete solution instead of a single step for each node.

\textbf{Iterative Improvement on Policy Model.}
We conduct experiments on LLaMA-3.1-8B-Instruct for one iterative update on MATH. Specifically, we collect the explored reasoning paths with the correct answers in the first round RL phase, and further use them for the second round supervised fine-tuning on the policy $\pi_{\text{SFT}}$ once more. We introduce ReST-MCTS* as a baseline, which uses iterative MCTS to collect training data for self-training. In Table~\ref{tab:self_improve}, we list the results of iterative updates on GSM8K and MATH datasets, demonstrating a continuous enhancement of the initial policy's performance. Notably, our framework achieves superior results compared to ReST-MCTS* (7.9\% vs. 1.7\% improvement on MATH), despite the latter employing an extra reward model to guide its searching.

\subsection{Ablation Studies}
We ablate core components of the RFTT: (1) \textit{RFTT w/o SFT Warmup} directly trains the original instruction model using ORM-based RL; (2) \textit{RFTT w/o MCTS} adopts random sampling and ORM in RL phase after SFT Warmup; (3) \textit{RFTT w/o PRM} employs MCTS sampling and ORM in RL phase after SFT Warmup.

\textbf{The Effectiveness of SFT Warmup.}
The SFT phase embeds the functional tokens into the model vocabulary, enabling self-improvement with token-guided exploration during the RL phase. To verify the effectiveness of our SFT phase, we conduct experiments to directly train the original instruction model without SFT warmup using pure RL. As compared from the first two lines of Table~\ref{tab:ablation}, even with the same random sampling method, the model warmuped with SFT (RFTT w/o MCTS) surpasses the pure RL model (RFTT w/o SFT warmup) by a large margin (5.7\% for Qwen-2.5-7B-instruct and 7.7\% for LLaMA-3.1-8B-instruct), indicating the token-guided reasoning can explore the reasoning space much more effectively after SFT initialization.

\begin{figure*}[!t]
    \centering
    \includegraphics[width=\linewidth]{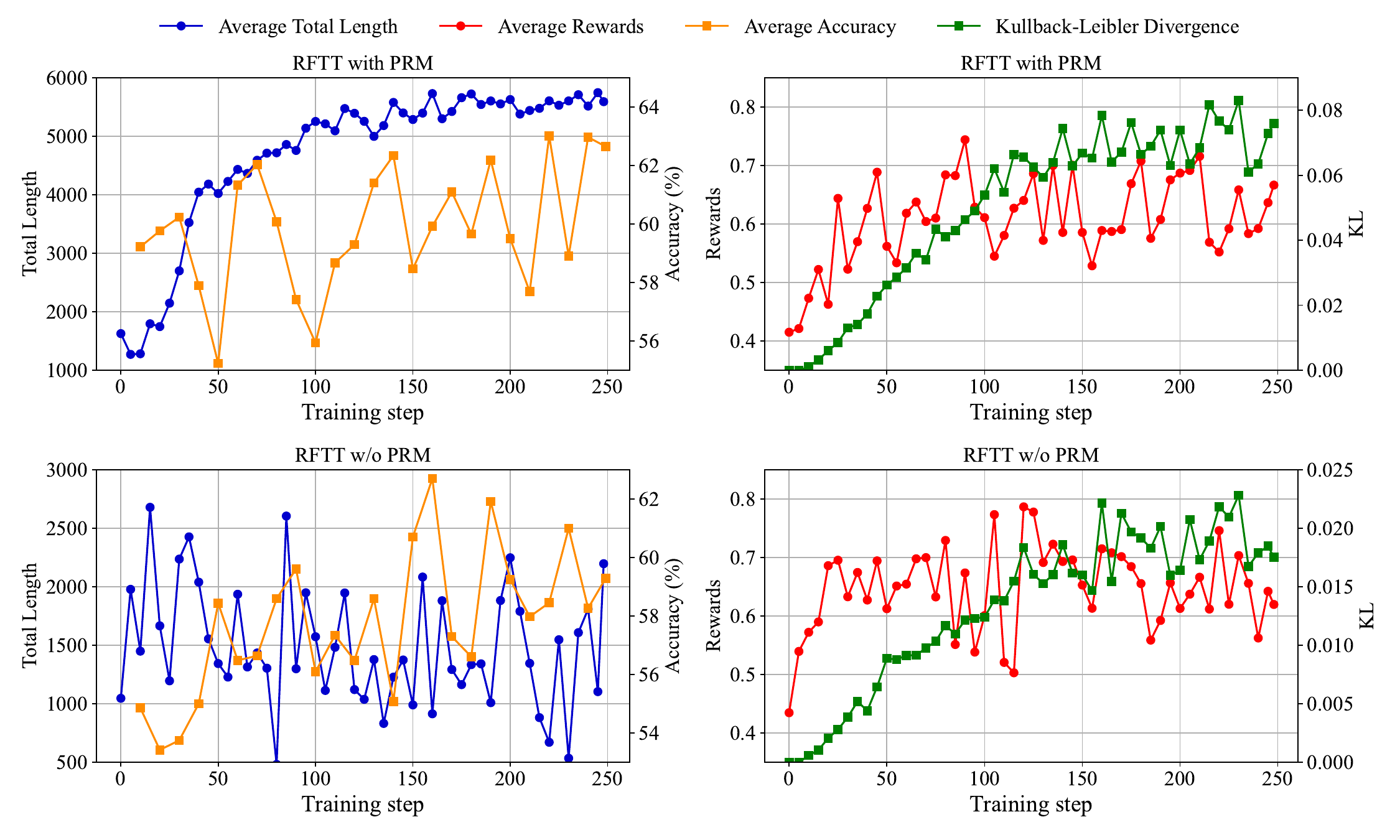}
    \vspace{-30pt}
    \caption{The training curve of RFTT with and w/o PRM on the training dataset during RL.}
    \vspace{-10pt}
    \label{fig:train_info}
\end{figure*}

\textbf{The Effectiveness of MCTS.} Our functional token-driven MCTS enables a more effective exploration of the solution space in both inference and training. We conducted an ablation study on MCTS sampling during the RL phase to investigate the influence of data diversity, as shown in Table~\ref{tab:ablation}. The outcomes reveal that random sampling in RL yields lower performance compared to sampling with MCTS, thereby underscoring the significance of data diversity in learning a better policy. The results show a 1.5\% performance gap between random sampling (mean 72.34) and MCTS-enhanced sampling (mean 73.92) after ORM-based RL across 5 benchmarks. Our final conclusion is that functional token-driven MCTS can explore diverse reasoning paths, providing a superior foundation for RL.

\textbf{The Effectiveness of PRM.}
As shown in Table~\ref{tab:ablation}, the results demonstrate that the use of PRM in RFTT consistently outperforms that of ORM (RFTT w/o PRM) by about 2\% on both LLaMA and Qwen on average. This might benefit from fine-grained rewards for intermediate reasoning steps, which mitigate error accumulation in long reasoning paths. Therefore the integration of PRM in MCTS is valuable for reasoning where outcome-based rewards fail to capture intermediate reasoning quality.

\subsection{Evolution Analysis}
\textbf{Response Length and Accuracy.}
We demonstrate the training curve of RFTT on Qwen-2.5-7B-Instruct in Fig.~\ref{fig:train_info}. As depicted, the average batch accuracies on the training set exhibit a consistent trend of improvement, while the trend of average response lengths displays distinct patterns under varying reward assignment strategies. The response length increases sharply within 50 training steps first and then gradually increases when using PRM in RFTT. The searching time of the policy model undergoes continuous improvement during the RL training process, developing the capacity for exploration and optimization with appropriate incentives. We observe that as the search depth increases, the policy reinforces predefined human-like reasoning behavior, including the self-correction mechanism and the exploration of diverse problem-solving solutions, thereby evolving to tackle challenging tasks. Please refer to Appendix~\ref{app:discuss} for more detailed discussions.

\textbf{Rewards and KL Divergence.}
We also demonstrate the development of average rewards and KL Divergence during the RL phase. Both of them improve steadily, which reflects the continuous improvement of the policy model. This phenomenon validates that through the appropriate definition of fine-grained reasoning steps (guided by functional tokens) and incentive mechanisms, it encourages the policy to autonomously explore advanced problem-solving strategies with high value in mathematical reasoning.

\section{Conclusions}
This work proposes RFTT, a reinforced fine-tuning framework that equips LLMs with self-play reasoning capabilities. By embedding learnable functional tokens into the model vocabulary, RFTT enables LLMs to internalize human-like reasoning behaviors, eliminating reliance on external prompts. Extensive experiments validate the effectiveness of RFTT, achieving significant performance gains and demonstrating scalability with inference-time search. This work advances the self-play paradigm by bridging token-guided reasoning and model training, offering a promising direction for developing resource-efficient, generalizable reasoning capabilities in smaller LLMs. Currently, our work focuses primarily on mathematical reasoning tasks; its effectiveness on broader reasoning domains needs further exploration. 

\section*{Acknowledgement}
This work was supported by the Alibaba Research Intern Program, the Hangzhou Joint Funds of the Zhejiang Provincial Natural Science Foundation of China under Grant No. LHZSD24F020001, and the Zhejiang Province High-Level Talents Special Support Program ``Leading Talent of Technological Innovation of Ten-Thousands Talents Program'' under Grant No. 2022R52046.




\bibliography{main}
\bibliographystyle{icml2025}

\newpage
\appendix
\onecolumn
\section{Functional Prompts\label{app:prompt}}

We present the designed functional prompts used in our experiments.

\definecolor{wkblue}{rgb}{0.2, 0.3, 0.6}
\definecolor{meta-color}{rgb}{0.5, 0.5, 0.5}
\definecolor{pinkcolor}{rgb}{1.0, 0.75, 0.8}  

\begin{tcolorbox}[colback=pinkcolor!10!white, colframe=pinkcolor!100!red, left=2mm, right=2mm, title=\small\centering\textcolor{black}{Prompt for Functional Token \texttt{<clarify>}}]
\begin{small}
You are an AI assistant to help me clarify questions by restating the original question and task in a clear and comprehensive manner. In your clarified version, ensure that all information from the original question is fully expressed. Following are some useful examples.\\

Original Question: Increasing the radius of a cylinder by $6$ units increased the volume by $y$ cubic units. Increasing the height of the cylinder by $6$ units also increases the volume by $y$ cubic units. If the original height is $2$, then what is the original radius?\\

Clarified Question: We are given a problem involving a cylinder where increasing the radius by 6 units and increasing the height by 6 units each results in an increase in volume by \( y \) cubic units. Our goal is to find the original radius of the cylinder, given that the original height is 2.\\

Original Question: A wooden model of a square pyramid has a base edge of 12 cm and an altitude of 8 cm. A cut is made parallel to the base of the pyramid that separates it into two pieces: a smaller pyramid and a frustum. Each base edge of the smaller pyramid is 6 cm and its altitude is 4 cm. How many cubic centimeters are in the volume of the frustum?\\

Clarified Question: Let’s understand what is being asked. Given a wooden model of a square pyramid has a base edge of 12 cm and an altitude of 8 cm. Also given a cut parallel to the base creates a smaller pyramid and a frustum. Note that the smaller pyramid has a base edge of 6 cm and an altitude of 4 cm. We need to find the volume of the frustum.\\

Question: \{ uesr\_question \}

\end{small}
\end{tcolorbox}

\begin{tcolorbox}[colback=pinkcolor!10!white, colframe=pinkcolor!100!red, left=2mm, right=2mm, title=\small\centering\textcolor{black}{Prompt for Functional Token \texttt{<analysis>}}]
\begin{small}
Given a math problem and an existing incomplete solution, your task is to provide a brief insight for the following problem. Note that it's not required to provide any specific solution process or calculation, but give a general guidance of sovling this problem. Usually one or two sentences are enough.\\

Question: \{ uesr\_question \}
\end{small}
\end{tcolorbox}

\begin{tcolorbox}[colback=pinkcolor!10!white, colframe=pinkcolor!100!red, left=2mm, right=2mm, title=\small\centering\textcolor{black}{Prompt for Functional Token \texttt{<subquestion>}}]
\begin{small}

Given a math problem and an existing incomplete solution, please propose a sub-problem that can be solved independently based on my current progress. The output format is limited to: "Let's ... now". Where ... indicates the omitted sub-problem that you should fill in. Note that do not propose sub-problems that have already been solved in previous steps.\\

Here is the input, please follow the restricted output format.\\

Question: \{ uesr\_question \}
\end{small}
\end{tcolorbox}

\begin{tcolorbox}[colback=pinkcolor!10!white, colframe=pinkcolor!100!red, left=2mm, right=2mm, title=\small\centering\textcolor{black}{Prompt for Functional Token \texttt{<next\_step>}}]
\begin{small}
Given a math problem and an existing incomplete solution, your task is to propose one next step in a smooth and proper way.\\

If no existing steps are provided, you need to briefly analyse the problem from scratch and then output the first step. Otherwise, you need to output the correct next step of the existing solution, following the ideas of the existing steps.\\

Your output should be a single reasoning step that may include reasoning, detailed calculations process, choosing answers, etc.\\

Question: \{ uesr\_question \}

\end{small}
\end{tcolorbox}

\begin{tcolorbox}[colback=pinkcolor!10!white, colframe=pinkcolor!100!red, left=2mm, right=2mm, title=\small\centering\textcolor{black}{Prompt for Functional Token \texttt{<direct\_answer>}}]
\begin{small}
Given a math problem and an existing incomplete solution, your task is to complete the remaining solution in a smooth and proper way. You need to give step-by-step solutions to the problem following the ideas of the existing steps, and do not repeat any existing steps.\\

Question: \{ uesr\_question \}

\end{small}
\end{tcolorbox}

\begin{tcolorbox}[colback=pinkcolor!10!white, colframe=pinkcolor!100!red, left=2mm, right=2mm, title=\small\centering\textcolor{black}{Prompt for Functional Token \texttt{<verify>}}]
\begin{small}
Given a problem and its previous solution steps, there is a correct subsequent solution and your subsequent solution. Please meticulously review your subsequent solution step by step and identify if any mistakes occur.

\# Correct solution\\
\{correct\_solution\}

\# Your subsequent solution\\
\{solution\}

\end{small}
\end{tcolorbox}

\section{Reasoning Cases\label{app:case}}
Our constructed data consists of diverse human-like reasoning behaviors, including double check, self-reflection, and self-correction. Following are some examples of diverse reasoning paths in mathematical reasoning.

\definecolor{wkblue}{rgb}{0.2, 0.3, 0.6}
\definecolor{meta-color}{rgb}{0.5, 0.5, 0.5}

\begin{tcolorbox}[breakable, colback=wkblue!10!white, colframe=wkblue!100!blue, left=2mm, right=2mm, title=\small\centering\textcolor{white}{Case 1: Self-Verification}]
\begin{small}

\# Problem: Complex numbers $a,$ $b,$ $c$ form an equilateral triangle with side length 18 in the complex plane.  If $|a + b + c| = 36,$ find $|ab + ac + bc|.$
\vspace{1.6mm}

\# Output: $\langle$clarify$\rangle$
The complex numbers \(a\), \(b\), and \(c\) form an equilateral triangle with side length 18 in the complex plane. Additionally, we know that \(|a + b + c| = 36\). Our task is to find the value of \(|ab + ac + bc|\).
$\langle$/clarify$\rangle$\\
$\langle$analysis$\rangle$
Use the properties of equilateral triangles in the complex plane, where the centroid (average) of the vertices is involved, and relate it to the given magnitude. Utilize the fact that for an equilateral triangle formed by complex numbers \(a\), \(b\), and \(c\), the expression \(a^2 + b^2 + c^2 = ab + ac + bc\).
$\langle$/analysis$\rangle$\\
$\langle$subquestion$\rangle$
First, Let me find the value of \(a + b + c\).
$\langle$/subquestion$\rangle$\\
$\langle$next\_step$\rangle$
Since the centroid of the equilateral triangle formed by \(a\), \(b\), and \(c\) is given by \(\frac{a + b + c}{3}\), and we know \(|a + b + c| = 36\), we can find \(|a + b + c|\) directly as \(36\). However, the next logical step is to use this information to find \(a + b + c\).

Given \(|a + b + c| = 36\), we can denote \(a + b + c = z\), where \(|z| = 36\). The specific process for this step is to recognize that \(a + b + c\) is a complex number with magnitude 36, but without loss of generality, we can assume \(a + b + c = 36e^{i\theta}\) for some angle \(\theta\). For simplicity, we can consider the case where \(\theta = 0\), so \(a + b + c = 36\). This assumption simplifies the calculation without affecting the final result due to the properties of magnitudes.

So, we have:
\[a + b + c = 36.\]
$\langle$/next\_step$\rangle$\\
$\langle$subquestion$\rangle$
Next, I need to find the value of \(|ab + ac + bc|\).
$\langle$/subquestion$\rangle$\\
$\langle$next\_step$\rangle$
Moreover, I will use the identity for the sum of products of roots of a polynomial.
\[
|ab + ac + bc| = \left| \frac{(a+b+c)^2 - (a^2 + b^2 + c^2)}{2} \right|.
\]

$\langle$/next\_step$\rangle$\\
$\langle$subquestion$\rangle$
Then, we must calculate \(a^2 + b^2 + c^2\).
$\langle$/subquestion$\rangle$\\
$\langle$next\_step$\rangle$
Using the property of equilateral triangles in the complex plane, we know that \(a^2 + b^2 + c^2 = ab + ac + bc\). However, we can also express \(a^2 + b^2 + c^2\) in terms of \(a + b + c\) and \(ab + ac + bc\):
\[
a^2 + b^2 + c^2 = (a + b + c)^2 - 2(ab + ac + bc).
\]
Given \(a + b + c = 36\), we substitute to get:
\[
a^2 + b^2 + c^2 = 36^2 - 2(ab + ac + bc).
\]
Now, let's solve for \(ab + ac + bc\).
$\langle$/next\_step$\rangle$\\
$\langle$subquestion$\rangle$
To continue, I need to solve for \(ab + ac + bc\).
$\langle$/subquestion$\rangle$\\
$\langle$next\_step$\rangle$
We already have:
\[
a^2 + b^2 + c^2 = 36^2 - 2(ab + ac + bc).
\]
And since \(a^2 + b^2 + c^2 = ab + ac + bc\), we substitute to get:
\[
ab + ac + bc = 36^2 - 2(ab + ac + bc).
\]
Rearrange to solve for \(ab + ac + bc\):
\[
3(ab + ac + bc) = 36^2,
\]
\[
ab + ac + bc = \frac{36^2}{3} = \frac{1296}{3} = 432.
\]
Now, let's find \(|ab + ac + bc|\).
$\langle$/next\_step$\rangle$\\
$\langle$verify$\rangle$
\textcolor{teal}{Hold on, let me confirm that.} The given steps correctly use the properties of equilateral triangles in the complex plane, including that the centroid (average) of the vertices is involved and the identity \(a^2 + b^2 + c^2 = ab + ac + bc\). The calculation for \(ab + ac + bc\) is correct: \[3(ab + ac + bc) = 36^2,\] \[ab + ac + bc = \frac{1296}{3} = 432.\] The magnitude of a real number is the number itself, so \(|ab + ac + bc| = 432\). There are no mistakes in the logic or calculations.
$\langle$/verify$\rangle$\\
$\langle$output$\rangle$
So the answer is \(\boxed{432}\).
$\langle$/output$\rangle$

\end{small}
\end{tcolorbox}

\begin{tcolorbox}[breakable, colback=wkblue!10!white, colframe=wkblue!100!blue, left=2mm, right=2mm, title=\small\centering\textcolor{white}{Case 2: Self-Reflection and Self-Correction}]
\begin{small}
\# Problem: Suppose that $f(x)=\frac{1}{2x+b}$. For what value of $b$ does $f^{-1}(x)=\frac{1-2x}{2x}$?"
\vspace{1.6mm}

\# Output: $\langle$analysis$\rangle$
To find the value of \(b\) for which the inverse function \(f^{-1}(x) = \frac{1 - 2x}{2x}\) matches with \(f(x) = \frac{1}{2x + b}\), we start by understanding that if \(y = f(x)\), then \(x = f^{-1}(y)\). This means substituting \(f(x)\) into the given form of \(f^{-1}(x)\) should yield \(x\).
$\langle$/analysis$\rangle$\\
$\langle$next\_step$\rangle$
First, substitute \(f(x) = \frac{1}{2x + b}\) into the equation for \(f^{-1}(x)\), giving us: \\\[x = \frac{1 - 2\left(\frac{1}{2x + b}\right)}{2\left(\frac{1}{2x + b}\right)}.\]\\This is based on the principle that applying a function and its inverse in succession returns the original input, i.e., \(f^{-1}(f(x)) = x\).
$\langle$/next\_step$\rangle$\\
$\langle$next\_step$\rangle$
Simplify the right-hand side:\\\[x = \frac{1 - \frac{2}{2x + b}}{\frac{2}{2x + b}} = \frac{(2x + b) - 2}{2} = \frac{2x + b - 2}{2}.\]

The simplification involves combining terms and recognizing that dividing by a fraction is equivalent to multiplying by its reciprocal.
$\langle$/next\_step$\rangle$\\
$\langle$next\_step$\rangle$
For the expression to equal \(x\), we set the simplified form equal to \(x\):\\\[x = \frac{2x + b - 2}{2}.\]

To solve for \(b\), we equate the numerators (since denominators are already the same and non-zero), leading to:\\\[2x = 2x + b - 2.\]
$\langle$/next\_step$\rangle$\\
$\langle$direct\_answer$\rangle$
From the equation \(2x = 2x + b - 2\), subtract \(2x\) from both sides to isolate \(b\):\\\[0 = b - 2,\]

which simplifies to:\\\[b = 2.\]\\Thus, the value of \(b\) that satisfies the condition is \(b = 2\), confirming the expected answer.
$\langle$/direct\_answer$\rangle$\\
$\langle$verify$\rangle$
But I have a feeling that it’s a little complex. Maybe I should \textcolor{teal}{consider a simpler approach.}
$\langle$/verify$\rangle$\\
$\langle$analysis$\rangle$
To find the value of \( b \), first express \( f^{-1}(x) \) in terms of \( f(x) \) and then solve for \( b \).
$\langle$/analysis$\rangle$\\
$\langle$subquestion$\rangle$
First, I should solve for \( f^{-1}(x) \) in terms of \( x \).
$\langle$/subquestion$\rangle$\\
$\langle$next\_step$\rangle$
Start with \( y = f(x) = \frac{1}{2x + b} \). Swap \( x \) and \( y \) to get \( x = \frac{1}{2y + b} \). Then solve for \( y \) to find \( f^{-1}(x) \).
$\langle$/next\_step$\rangle$\\
$\langle$subquestion$\rangle$
Next, I need to solve for \( y \) in the equation \( x = \frac{1}{2y + b} \).
$\langle$/subquestion$\rangle$\\
$\langle$next\_step$\rangle$
Multiply both sides by \( 2y + b \) to get rid of the denominator, then isolate \( y \).
$\langle$/next\_step$\rangle$\\
$\langle$next\_step$\rangle$
Multiply both sides by \( 2y + b \) to obtain \( x(2y + b) = 1 \). Then distribute \( x \) on the left side to get \( 2xy + bx = 1 \).
$\langle$/next\_step$\rangle$\\
$\langle$direct\_answer$\rangle$
Next, we will isolate \( y \).

Starting from the equation \( 2xy + bx = 1 \):\\\[ 2xy + bx = 1 \]\\Subtract \( 2xy \) from both sides:\[ bx = 1 - 2xy \]\\Divide both sides by \( b \):

\textcolor{red}{\[ y = \frac{1 - 2xy}{b} \]}\\Since we know that \( f^{-1}(x) = \frac{1 - 2x}{2x} \), we can equate the expressions for \( y \):\[ \frac{1 - 2xy}{b} = \frac{1 - 2x}{2x} \]\\For the above equality to hold for all \( x \), the numerators and denominators must match. Therefore, we can compare the denominators directly:\\\[ b = 2x \cdot 2x = 2 \cdot 2 = 4 \]
$\langle$/direct\_answer$\rangle$\\
$\langle$verify$\rangle$
Oops, I seem to get two different results. Actually, I'm getting confused. \textcolor{blue}{I think I should review carefully} about my previous solutions. 

The mistake occurs when dividing by \( b \) and incorrectly expressing \( y \) as \( y = \frac{1 - 2xy}{b} \).

The correct form should be \( y = \frac{1 - bx}{2x} \).

As a result, the value of \( b \) is incorrectly determined.
$\langle$/verify$\rangle$\\
$\langle$refine$\rangle$
Correct the expression for \( y \) after isolating it: \( y = \frac{1 - bx}{2x} \).

Equate this to the given \( f^{-1}(x) = \frac{1 - 2x}{2x} \):

\[ \frac{1 - bx}{2x} = \frac{1 - 2x}{2x} \]

Since the denominators are the same, equate the numerators:

\[ 1 - bx = 1 - 2x \]

- Subtract 1 from both sides:

\[ -bx = -2x \]

- Divide both sides by \(-x\) (assuming \( x \neq 0 \)):

\[ b = 2 \]

Thus, the value of \( b \) is \( 2 \).
$\langle$/refine$\rangle$\\
$\langle$output$\rangle$
The answer is: \(\boxed{2}\).
$\langle$/output$\rangle$
\end{small}
\end{tcolorbox}

\section{Detailed MCTS\label{app:mcts}}
The MCTS algorithm is capable of balancing exploration and exploitation within vast environmental spaces, enabling it to quickly identify high-quality steps or trajectories. Due to its focus on exploring unseen states and sampling high-quality trajectories, which is a key aspect of online sampling in reinforcement learning, MCTS has been widely applied in reinforcement learning tasks, such as AlphaGo and AlphaZero~\citep{silver2018general}. The standard MCTS algorithm consists of four stages: \textit{selection}, \textit{expansion}, \textit{simulation}, and \textit{backpropagation}.

(1) \textbf{Selection.} Starting from the root node of the tree, the child node is selected downward based on the UCT formula, until an \textit{unexpanded} node is reached. The UCT selection can be described as follows:
\begin{equation}
    \texttt{UCT}(s) = \frac{Q(s)}{N(s)} + c\cdot\sqrt{\frac{\ln \tilde{N}(s) }{N(s)} },
\end{equation}
\begin{equation}
a_t = 
    \begin{cases}
        \text{max}_{a \in \mathcal{A}} ~\pi_\theta(a|s_{0:t}), & \exists ~a \in U(s_{t}), \\
        \text{max}_{a \in \mathcal{A}} ~\texttt{UCT}(s_{t}),     & \text{otherwise}, \\
    \end{cases}
\end{equation}
where $c$ serves as an exploration weight that encourages the exploration of less-visited nodes when given a higher value; $N(s)$ indicates the number of times the node at state $s$ has been visited; $\mathcal{A}$ represents the set of possible actions that can be selected from the current state $s_t$; $U(s_{t})$ denotes unexplored functional tokens of intermediate reasoning step $s_{t}$; and $\pi_\theta(a|s_{0:t})$ is the probability to sample functional token $a$ based on the current state.

In LLM  sampling, each action in MCTS corresponds to a reasoning step. The sequence of generated steps forms the current state. Selecting an action in MCTS means generating the next step, and the next state is obtained by appending this step to the already generated steps. This state transition is deterministic, so we can simplify $Q(s_t, a_{t+1})$ as $Q(s_{t+1})$ and store it at the $s_{t+1}$ node.

(2) \textbf{Expansion.} The selected node is expanded by choosing an action that has not yet been selected, thereby generating the next state node. Notably, the selection of certain actions has dependency relationships. For example, \texttt{<clarify>} ($a_1$) can only be expanded after the root question, and \texttt{<subquestion>} ($a_3$) can only expand the \texttt{<next\_step>} ($a_4$). All prerequisite dependencies are presented in Table~\ref{tab:prerequisite}.

(3) \textbf{Simulation.} From the next state node entered after Expansion, one or more simulations are performed. A simulation involves randomly selecting an action from the current node, generating the next state node, and continuing this process until a terminal state is reached.

(4) \textbf{Backpropagation.} Based on the results of the Simulation, Backpropagation updates the $Q$ and $N$ values for all nodes along the path from the current node to the root node. The update formula is as follows:
\begin{equation}
    N(s_t) \leftarrow N(s_t) + 1,
\end{equation}
\begin{equation}
    Q(s_t) \leftarrow (Q(s_t) + r) / N(s_t),
\end{equation}
where $r$ is the terminal reward obtained from the simulation reaching the terminal node. In the $Q$-update formula, $N(s_t)$ refers to the updated visit count for the node at state $s_t$.

\begin{table}[!t]
\begin{minipage}{0.5\textwidth}
\caption{The detailed information of five datasets.}
\label{tab:data_detail}
\vspace{6pt}
\begin{center}
\begin{small}
\setlength{\tabcolsep}{6pt}
\renewcommand{\arraystretch}{1.2}
\begin{tabular}{@{\extracolsep{\fill}}crrc}
\toprule
\textbf{Dataset} & \textbf{\# Train} & \textbf{\# Test} & \textbf{\# Domain}\\
\midrule
MATH             & 7500    & 5000 & In-Domain \\
\midrule
GSM8K            & 7473    & 1319 & Out-of-Domain \\
\midrule
SVAMP            & -    & 1000 & Out-of-Domain \\
\midrule
Olympiad Bench   & -    & 675 & Out-of-Domain \\
\midrule
AMC              & -    & 40 & Out-of-Domain \\
\bottomrule
\end{tabular}
\end{small}
\end{center}
\end{minipage}
\hfill
\begin{minipage}{0.5\textwidth}
\caption{The dependencies between actions.}
\vspace{2pt}
\label{tab:prerequisite}
\begin{center}
\begin{small}
\setlength{\tabcolsep}{6pt}
\renewcommand{\arraystretch}{0.8}
\begin{tabular}{@{\extracolsep{\fill}}cl}
\toprule
\textbf{Previous Action} & \multicolumn{1}{c}{\textbf{Next Action}} \\
\midrule
clarify ($a_1$)      & $a_2, a_3, a_4, a_5$  \\
\midrule
analysis ($a_2$)     & $a_3, a_4, a_5$ \\
\midrule
subquestion ($a_3$)  & $a_4$  \\
\midrule
next\_step ($a_4$)   & $a_3, a_4, a_5, a_6, a_8$  \\
\midrule
direct\_answer ($a_5$) & $a_6, a_8$  \\
\midrule
verify ($a_6$)       & $a_3, a_4, a_5, a_6, a_7, a_8$ \\
\midrule
refine ($a_7$)      & $a_3, a_4, a_5, a_6, a_8$ \\
\bottomrule
\end{tabular}
\end{small}
\end{center}
\end{minipage}
\end{table}

\section{Experimental Detials\label{app:exp}}
\begin{figure}[!b]
    \centering
    \includegraphics[width=1\linewidth]{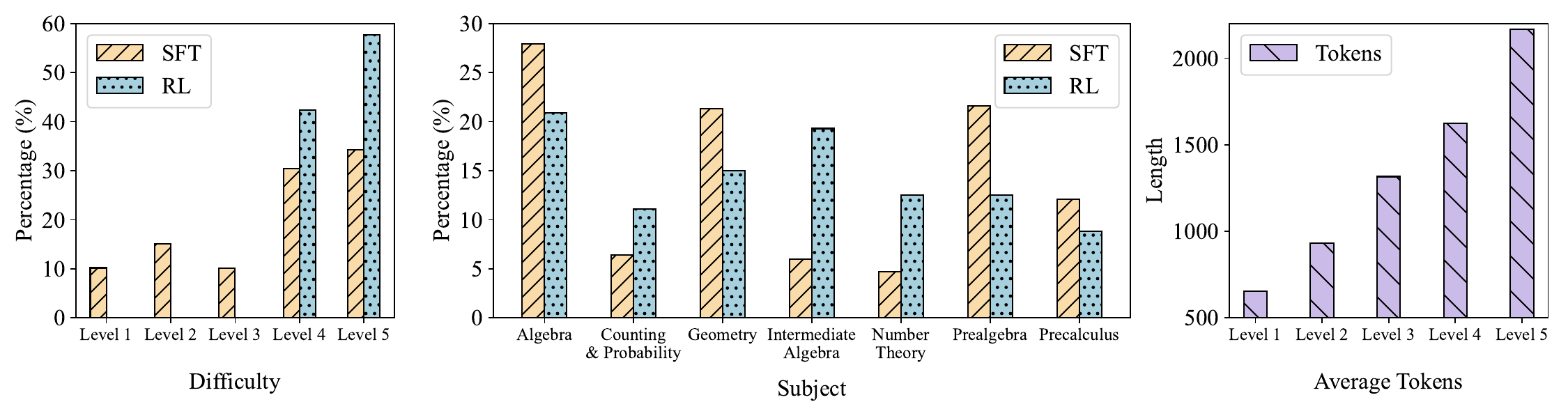}
    \vspace{-25pt}
    \caption{Detailed statistical information of training dataset in two-phase training.}
    \vspace{-10pt}
    \label{fig:dataset_details}
\end{figure}

We list detailed information of our training and evaluation dataset in Table~\ref{tab:data_detail}. Specifically, we select 1000 and 3994 questions from the training set of MATH for SFT and RL respectively. Recent studies have revealed that focusing on more difficult problems can better enhance reasoning capability~\cite{min2024imitateexploreselfimprovereproduction, team2025kimi}, thus we increase the proportion of difficult questions (Level 4 and Level 5) in our training. The data distribution of SFT and RL phases are shown in Fig.~\ref{fig:dataset_details}.

In evaluating the performance of superior LLMs, we take the relevant metrics from their official technical reports~\cite{achiam2023gpt, anthropic2024claude, liu2024deepseek, dubey2024llama, yang2024qwen25}. For non-reported metrics, we report their accuracies on corresponding datasets using official API with zero-shot (original question without any prompts). For the results of CoT, we demonstrate the CoT prompt template used to evaluate in Table~\ref{tab:cot_template}.

\begin{table}[t]
\caption{CoT template for evaluation. The variable \textcolor{red}{question} will be replaced with the specific problem when testing.}
\vspace{2pt}
\centering
\small
\begin{tabular}{l}
\toprule
You are a mathematics expert that solves problems step by step.\\
Note that you should output your final answer with the format of ``The answer is: \texttt{boxed\{<ANSWER>\}}.", where \texttt{<ANSWER>} should \\ be a numeric result or a math expression.\\
Question: \{\textcolor{red}{question}\}\\
Let's think step by step.\\
 \bottomrule
\end{tabular}
\vspace{-0.3cm}
\label{tab:cot_template}
\end{table}

\section{Further Discussion\label{app:discuss}}
\subsection{How do Functional Tokens Benefit?}
There are usually two challenges when using MCTS in training~\citep{guo2025deepseek}: (1) Unlike traditional games that have well-defined action space, the combinatorial vocabulary space leads to an exponential growth in candidate reasoning paths; (2) Training a fine-grained value model to guide searching is quite difficult.

The introduction of functional tokens with rich semantics can diversify node exploration while constraining the search space. Notably, we can use the intrinsic probability of functional tokens which adjust dynamically during training to guide its search process, thereby estimating the need for training an extra value model. Furthermore, the functional tokens naturally divide the reasoning process into a series of fine-grain steps, which can seamlessly integrate with PRM.

\subsection{Is SFT Warmup with Functional Tokens Needed?}
Recently Deepseek R1-Zero~\cite{guo2025deepseek} has demonstrated the emergence of self-reflection in the LLM's response during the pure RL phase. We also encounter the occurrence of this kind of feature when conducting pure RL on Qwen-2.5-7B-Instruct (RFTT w/o SFT Warmup), as shown in Table~\ref{tab:slef-reflection}. But such a pattern occurs at an extremely low rate (about 1.5\%) on Qwen-2.5-7B-Instruct and there is even no such case observed on LLaMA-3.1-8B-Instruct. The power of RL is to reinforce high-quality reasoning patterns like self-reflection and exploration of different approaches, thus maximizing expected rewards by leveraging increased test-time computation. However, in a model with smaller parameters, the limitation arises from their inherent preference for syntactic patterns established in pretraining, which further restricts the reinforcement of sophisticated reasoning patterns. Therefore, SFT warmup with functional tokens can play important roles in faster convergence and stable training.

\begin{table}[t]
\caption{The occurrence of self-reflection and self-correction during pure RL on Qwen-2.5-7B-Instruct.}
\vspace{2pt}
\centering
\begin{tabular}{l}
\toprule 
Question: In $\triangle{RST}$, shown, $\sin{R}=\frac{2}{5}$. What is $\sin{T}$? [asy]...[/asy]\\
\midrule
Response: Given that $\triangle RST$ is a right triangle with the right angle at $S$, we know that $\sin R = \frac{\text{opposite}}{\text{hypotenuse}} = \frac{RT}{RS}$. Here,\\$RT = \sqrt{21}$ and $RS = 5$. Therefore, we have:$\sin R = \frac{\sqrt{21}}{5} = \frac{2}{5}$. This implies that $\sqrt{21} = 2$, which is \textcolor{red}{incorrect} based\\ on the given information. \textcolor{red}{Instead}, we should use the correct value of \(\sin R = \frac{2}{5}\) and find the correct lengths. Let's\\ denote the length of \(RT\) as \(a\) and the length of \(RS\) as \(c = 5\). Then, we have: $\sin R = \frac{a}{c} = \frac{a}{5} = \frac{2}{5} \implies a = 2$. Now,\\ we can find the length of \(ST\) using the Pythagorean theorem:$ST = \sqrt{RS^2 - RT^2} = \sqrt{5^2 - 2^2} = \sqrt{25 - 4} = \sqrt{21}$.\\In \(\triangle RST\), $\sin T = \frac{\text{opposite}}{\text{hypotenuse}} = \frac{RS}{RT} = \frac{5}{2}$. \textcolor{red}{However, we need to re-evaluate} the correct trigonometric relationship. \\Since $\sin R = \frac{2}{5}$, we know that $\cos R = \sqrt{1 - \sin^2 R} = \sqrt{1 - \left(\frac{2}{5}\right)^2} = \sqrt{1 - \frac{4}{25}} = \sqrt{\frac{21}{25}} = \frac{\sqrt{21}}{5}$. In a right triangle,\\ \(\sin T = \cos R\). Therefore: $\sin T = \cos R = \frac{\sqrt{21}}{5}$.\\Thus, the value of \(\sin T\) is: $\boxed{\frac{\sqrt{21}}{5}}$. \\
\hline
\end{tabular}
\label{tab:slef-reflection}
\end{table}

\subsection{Discussion of Response Lengths.}
\begin{figure}
    \centering
    \includegraphics[width=1\linewidth]{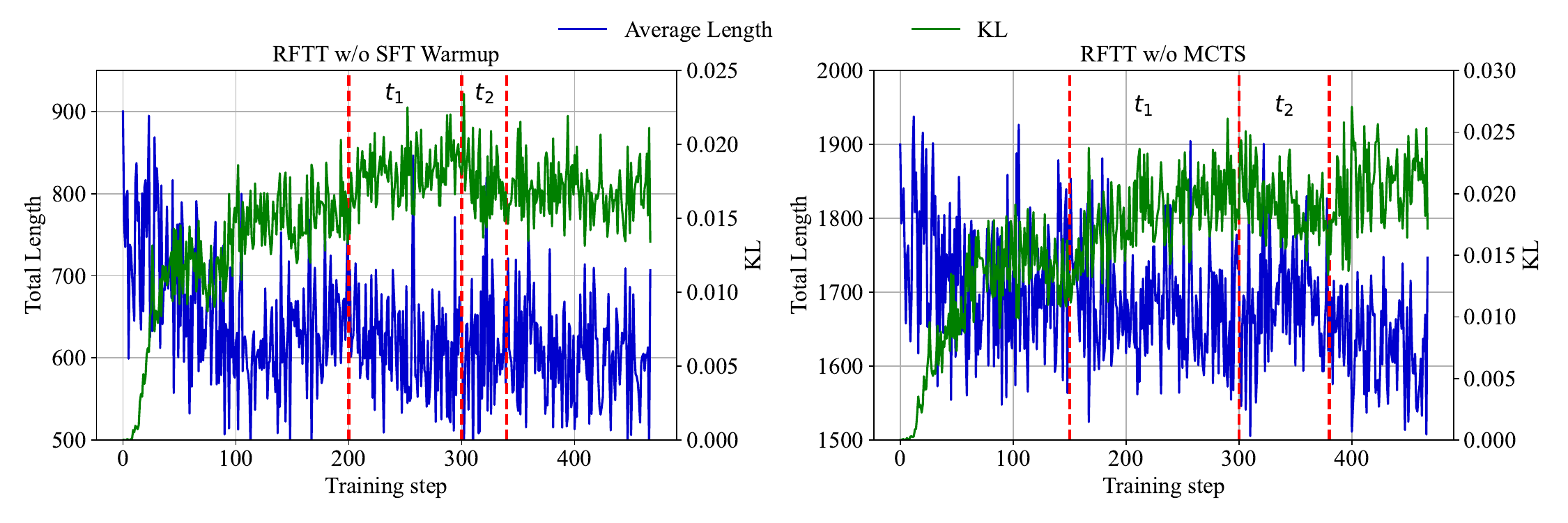}
    \vspace{-20pt}
    \caption{The training curve of RFTT w/o SFT Warmup and w/o MCTS on the training dataset during RL.}
    \label{fig:purerl_len_kl}
\end{figure}

Recent research suggests that the reasoning capabilities of the model may benefit from longer response lengths~\citep{guo2025deepseek}.  In Fig.~\ref{fig:purerl_len_kl}, we present the training curve of RFTT w/o SFT Warmup and RFTT w/o MCTS on LLaMA-3.2-3B-Instruct as an example. We conduct RFTT w/o SFT Warmup with the prompt template in Table~\ref{tab:cot_template}. Surprisingly we find that after an initial drop in the response lengths of the policy model, the lengths rebound but persistently remain below the response lengths observed in the original pretrained model. Moreover, we observe that as the response lengths drop, the KL divergence between the current policy and the initial model increases during training period $t_1$; while the lengths increase, the KL divergence drops during training period $t_2$. This indicates a potential correlation between response lengths and KL constraints. It seems that the subsequent increase in response lengths cannot be attributed to the exploration of sophisticated behaviors, but rather primarily stems from the impact of the KL penalty.

In contrast, the response lengths of RFTT w/o PRM in Fig.~\ref{fig:train_info} do not exhibit a downward trend. This is likely due to the benefits of our functional token-guided tree search, which provides a diverse exploration space. Furthermore, the use of PRM in MCTS introduces fine-grained reward assignments in intermediate functional tokens. Such mechanism significantly mitigates the exploration limitations imposed by the KL divergence penalty, potentially elucidating the observed length expansion in RFTT with PRM during the training process. This highlights the effectiveness of RFTT, where the functional tokens serve as a bridge that seamlessly combines the employment of MCTS and PRM for learning to reason.

\end{document}